\documentclass[letterpaper, 10 pt, conference]{ieeeconf}  

\IEEEoverridecommandlockouts                              

\usepackage{ifpdf}
\usepackage{graphicx}
\ifpdf
\usepackage{amsmath}
\usepackage{autobreak}
\usepackage{hhline}
\usepackage{multirow}
\usepackage{hyperref}
\usepackage{graphicx}

\usepackage{algorithmic}
\usepackage{algorithm}  
\usepackage{dblfloatfix}
\usepackage{blindtext}

\usepackage{amsmath}

\graphicspath{{Figs/}{Figs/PDF/}}
\else
\graphicspath{{Figs/}}
\fi
\usepackage{hhline}
\usepackage{color}
\usepackage{calc}
\let\labelindent\relax
\usepackage{enumitem}
\usepackage{bm,amsmath}

\usepackage{nomencl}
\usepackage{multirow}

\usepackage{url}
\usepackage{nomencl}
\usepackage{multirow}
\usepackage{multicol}
\usepackage{multirow}

\usepackage{bbding}
\usepackage{booktabs}
\usepackage{makecell}

\hypersetup{
	colorlinks = true,
	linkcolor = blue,
	citecolor = blue,
    urlcolor = blue
}
\usepackage{subcaption}

\usepackage{xcolor}

\usepackage{caption}
\usepackage[flushleft]{threeparttable}
\usepackage{cite}
\usepackage{times}
\usepackage{epsfig}
\usepackage{graphicx}
\usepackage{amsmath}
\usepackage{amssymb}

\usepackage[maxfloats=256]{morefloats}
\title{\LARGE \bf
Learn from the Past: Language-conditioned Object Rearrangement with Large Language Models
}

\author{Guanqun Cao, Ryan Mckenna, Erich Graf and John Oyekan
\thanks{This work was funded by the EPSRC project ``A digital COgnitive architecture to achieve Rapid Task programming and flEXibility in manufacturing robots through human demonstrations (DIGI-CORTEX)'' (EP/W014688/2).}
\thanks{Cao, Mckenna and Oyekan are with the Department of Computer Science, University of York, York, YO10 5DD, United Kingdom. Graf is with the Department of Psychology, University of Southampton, Southampton SO17 1PS. Emails:~\{\tt\small guanqun.cao, ryan.mckenna, john.oyekan\}@york.ac.uk. 
}
}

\begin{document}

\maketitle
\thispagestyle{empty}
\pagestyle{empty}

\begin{abstract}

Object manipulation for rearrangement into a specific goal state is a significant task for collaborative robots. 
Accurately determining object placement is a key challenge, as misalignment can increase task complexity and the risk of collisions, affecting the efficiency of the rearrangement process.
Most current methods heavily rely on pre-collected datasets to train the model for predicting the goal position. As a result, these methods are restricted to specific instructions, which limits their broader applicability and generalisation.
In this paper, we propose a framework of flexible language-conditioned object rearrangement based on the Large Language Model (LLM).
Our approach mimics human reasoning by making use of successful past experiences as a reference to infer the best strategies to achieve a current desired goal position.
Based on LLM's strong natural language comprehension and inference ability, our method generalises to handle various everyday objects and free-form language instructions in a zero-shot manner.
Experimental results demonstrate that our methods can effectively execute the robotic rearrangement tasks, even those involving long sequences of orders.

\end{abstract}

\section{INTRODUCTION}

Robots have the potential to assist humans with various daily tasks, such as placing objects in desired locations, tidying up toys by gathering them into a designated container and performing household cleaning. 
These tasks can be categorised as object rearrangement, in which robots manipulate objects to achieve a specified goal state within a physical environment~\cite{batra2020rearrangement}. 
Effective object rearrangement involves several sub-tasks, including the recognition of object states within the environment, the inference of difference between the current and goal state, and the manipulation of objects accordingly.
While there are many advanced solutions for object recognition and manipulation in computer vision, tactile sensing and robotic fields \cite{li2022tactile}\cite{cao2024multimodal}, we argue that there is a lack of intelligent and flexible methods that can perform human-like thinking to infer optimal object placement in rearrangement tasks while generalising to unforeseen scenarios not included in the training dataset. This is especially true for \textbf{intra-class} tasks. Currently, most existing methods attempt to learn a mapping from the current position to the goal position~\cite{mees2017metric}\cite{yuan2022sornet}. 

In these methods, a large dataset describing the goal position for the rearrangement is collected and used to train deep learning algorithms running on the robot. Using the current position as input, the algorithm predicts the goal position and directs the robot's movement accordingly. However, these methods rely heavily on the training dataset and makes them learn only specific patterns with specific objects and human instructions from the dataset. This works for \textbf{inter-class} tasks but when faced with a new environment containing different objects not included in the dataset or new instructions (\textbf{intra-class}), these methods do not generalise effectively \cite{vitiello2023one}. Most recently, generative models like DALL-E, which is pre-trained on web-scale data, have been used to generate the goal poses in rearrangement to improve the generalisation ability~\cite{kapelyukh2023dall}. However, these models can sometimes generate hallucinations, producing random objects and positions that require filtering in a further step.

\begin{figure}[t]
	\centering
	\includegraphics[trim=0 0 0 0, clip, width=0.85\columnwidth]{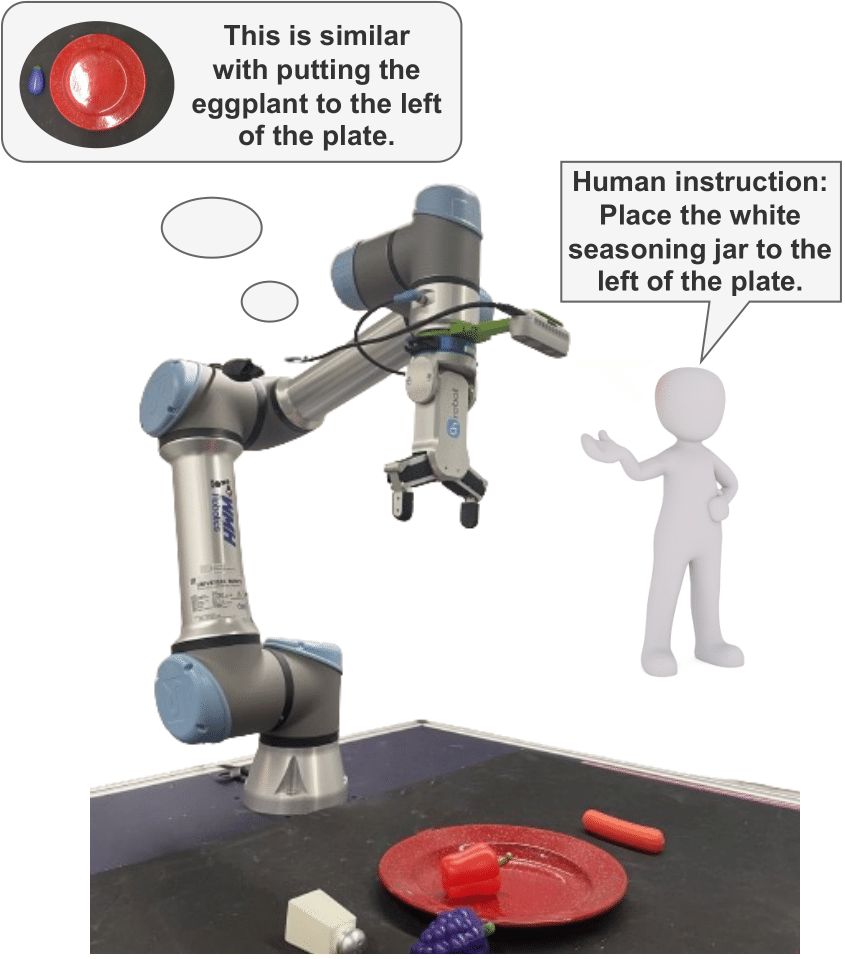}
	\caption{\textbf{\textit{Learn from the past.}} In our framework, the robot retrieves past experiences to find the most similar arrangement based on human instructions. By referencing previously successful arrangements, the robot can mimic human-like reasoning (See Fig. \ref{Architecture}), allowing it to infer the goal position for rearrangement more effectively.  
 }
	\label{fig:first}
\vspace{-1.5em}
\end{figure}

\begin{figure*}[t]
	\centering
	\includegraphics[trim=0 0 0 0, clip, width=1.75\columnwidth]{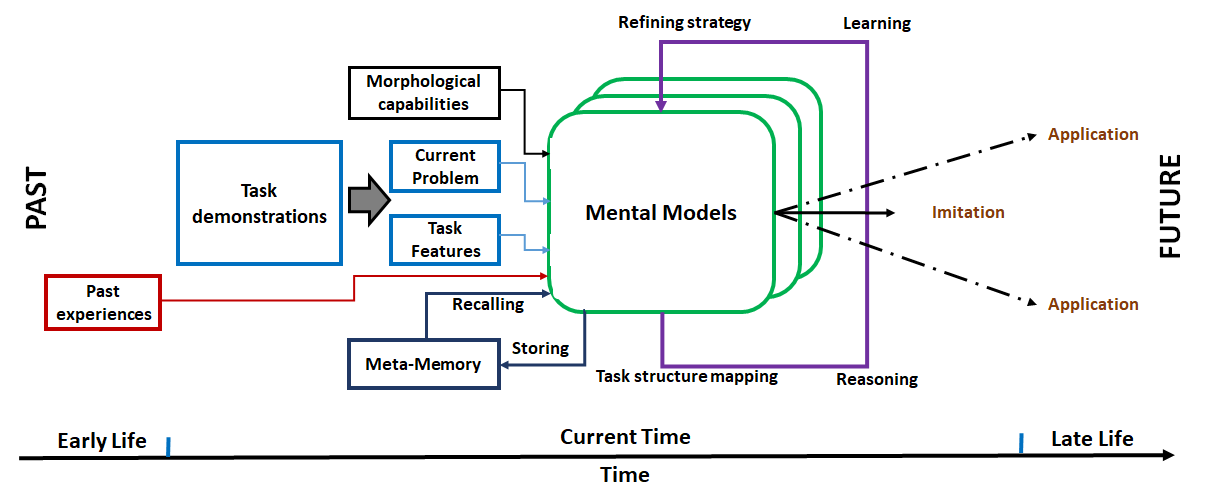}
	\caption{\textbf{\textit{Learn from the past in humans.}} A diagram showing how humans make use of past experiences to guide successful current and future task completions in life long learning \cite{barker1998mental}. Mental models are used as templates or references for current tasks. Successful mental models are stored for future reuse.
 }
	\label{Architecture}
\vspace{-1em}
\end{figure*}

As a result, achieving human's ability to adapt to changing circumstances and guide flexible behaviour \cite{klein2022medial}\cite{fine2022whole} through the use of similar past experiences (See Fig. \ref{Architecture}), still needs to be researched. When humans respond to new instructions to manipulate or rearrange items both in inter-task and intra-task scenarios, it is natural for them to draw on their past successful experiences~\cite{pally1997memory}\cite{barker1998mental}. 
For example, the instruction ``\textit{\textbf{place an apple on a plate}}" may remind them of a similar prior arrangement and instruction, such as ``\textit{\textbf{put an orange in a bowl}}". 
In this case, the experience of placing an orange in a bowl can serve as a \textbf{mental model template} that provides scaffolding for reasoning about the current task as well as generating the required trajectories and positions for completing the task involving both the apple and the plate. 
By leveraging past experiences, the completion efficiency of tasks is improved, as people distil and transfer knowledge from their previous experiences to new tasks (See Fig. \ref{Architecture}).

Following this inspiration, we introduce a framework that links past experiences with new instructions by leveraging natural language comprehension and reasoning (See Fig. \ref{fig:first}). Specifically, we make use of Large Language Models (LLMs) to draw on similar past experiences towards constructing templates for performing spatial reasoning in current rearrangement tasks. As a result, our proposed method allows the robot to mimic human-like flexibility in language conditioned (i.e., human language instructed) object rearrangement tasks while executing the tasks efficiently and accurately, even in tasks with long sequential orders.


The contributions of this paper are summarised as follows:
\begin{enumerate}
\item We propose a novel framework that leverages past successful experiences to address the problem of flexible object rearrangement tasks.
\item Our framework is not limited by a training dataset and can be generalised to various tasks using free-form natural language instructions in a zero-shot manner.
\item By emulating human reasoning in leveraging past experiences, our framework significantly enhances the performance of current rearrangement tasks.
\end{enumerate}


\begin{figure*}[t]
	\centering
	\includegraphics[trim= 0 0 0 0, clip, width=2\columnwidth]{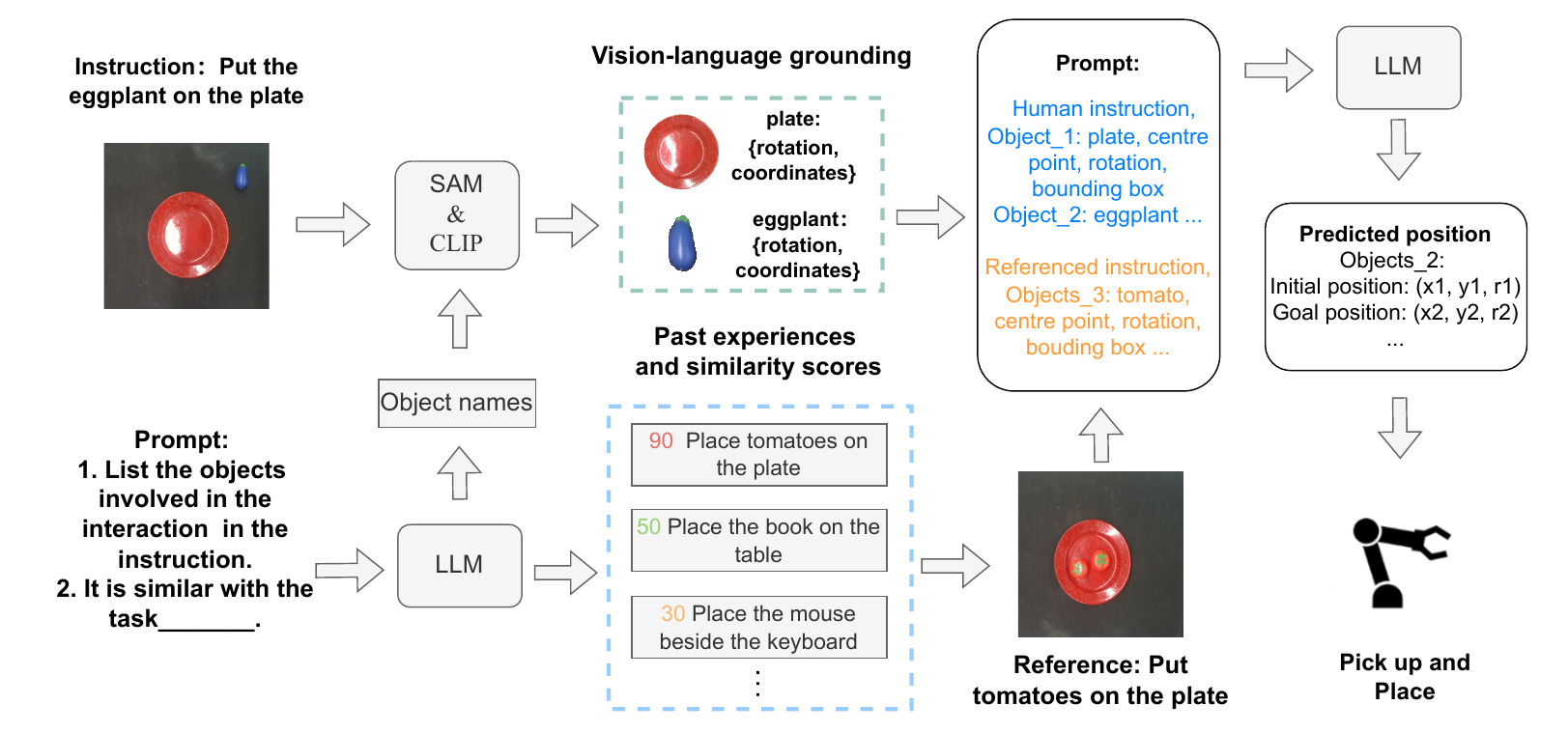}
	\caption{\textbf{Illustration of the proposed framework.} The robot uses SAM for visual perception and CLIP for semantic understanding to identify \textit{where} and \textit{what} objects are in the environment. The LLM then associate the most similar past experience with instruction and uses this similar experience as a template and reference. Finally, a prompt is created with spatial and semantic information, allowing the LLM to reason and predict the goal position for rearrangement.
	} 

	\label{fig:framework}
 \vspace{-1em}
\end{figure*}

\section{Related works}\label{relatedwork} 
In this section, we will first review works on predicting the goal positions in object rearrangement, followed by discussions on LLM's applications in robotics. Finally, we will review the Visual Language Model (VLM) which connects visual perception and language instruction in robotics.
\subsection{Prediction of Goal Position in Rearrangement}
The prediction of the goal position (that is the positions of each physical object in the environment) plays an important role in the object rearrangement task. 
The goal position can be represented in various ways such as through logical predicates, visual images or natural language instructions \cite{batra2020rearrangement}.
In \cite{mees2017metric}\cite{yuan2022sornet}, the spatial relations of the logical predicates, such as \textit{inside}, \textit{on} and \textit{next to} are modelled between different objects to predict the goal position. Additionally, some prior works treat position prediction as a searching task by selecting from a set of sampled poses using graphs. 
For example, in \cite{lin2022efficient}, a GNN is trained to understand the rules of the task, with its nodes representing entities relevant to the task such as objects and goals, which are used to select an appropriate position.
In~\cite{sarch2022tidee}, an embodied agent is developed to detect out-of-place objects and infer plausible receptacles through a neural graph memory.
However, the methods mentioned above typically require a large dataset that is manually collected for learning purposes, which can be both costly and time-consuming. The approaches above can be combined with visual image detection and representation in which objects are manually arranged into goal states according to human preferences. In these situations, robots learn the latent space of user preferences from the image and manipulate the objects accordingly \cite{kang2018automated}\cite{kapelyukh2022my}\cite{ma2023learning}\cite{ma2024applying}.

Nevertheless, compared to only visual goal specification, language enriches human robot collaboration by offering a more intuitive way to describe the goal state. For example, in \cite{krishna2017visual}\cite{shridhar2020ingress}, researchers used vision to detect objects within the environment and established spatial as well as semantic relationships between them. Then robots were able to relocate a specified object with respect to a designated anchor object towards fulfilling a set of given natural language instructions. However, these methods concentrate on learning visual-semantic relationships and are unable to infer the goal position in a manner similar to human reasoning. 
In our framework, we use LLMs to draw on similar past successful experiences and to perform spatial reasoning in language-conditioned flexible rearrangement of objects. By providing sufficient information regarding the current task and similar previous tasks, including the object's name, position, etc., the robot is able to construct the appropriate template (or mental model in humans) for comprehension, reasoning and completing the current task. According to our current knowledge, this is the first of its kind.

\subsection{LLM and VLM in Language-conditioned Manipulation}
Recently, LLMs have become an active area of research in the field of robotics due to their extensive internalised knowledge and chain of thought generation~\cite{li2024foundation}. 
For example, in~\cite{brohan2023can}\cite{sarch2022tidee}\cite{huang2022language}, language models are applied to decompose instructions into sequences of sub-steps. Based on the observed environment, the robot can execute the feasible sub-tasks to successfully complete the instructions. 
Contrary to decomposing instructions, in~\cite{wu2023tidybot}, a LLM is applied to summarise the rules of arrangement from a small number of examples by leveraging its few-shot learning capabilities.
Additionally, in~\cite{zhang2023bootstrap}, the proposed method refines long-horizon behaviours by chaining basic actions together with guidance from an LLM.
In~\cite{tziafas2024lifelong}, a wake-sleep framework is applied, where the wake phase employs an LLM-based actor-critic to interact with the environment based on human demonstrations and hints, and the sleep phase clusters experiences with an LLM abstractor. 
However, these methods focus on reordering basic actions to form new behaviors rather than learning entirely novel skills.
In our framework,  we employ a LLM to connect past arrangements with new instructions and use it as a reference (or mental model)  for spatial reasoning in estimating a more accurate target state, for the first time.


Most lately, Vision-Language Models (VLMs) have become a successful paradigm for aligning visual information with language, enabling robots to perceive the world in a multimodal manner.
In~\cite{zitkovich2023rt}, an end-to-end model based on VLMs is proposed. This model learns from large amounts of online data to create action commands that demonstrate robust generalisation capabilities in unseen environments. 
In~\cite{driess2023palm}, observations like images and state estimates are integrated into the language embedding space, facilitating more efficient inferences, particularly for sequential decision-making. 
However, these methods usually use large robotic data for training, which is a bottleneck because of limited data availability.
To address this issue, \cite{huang2023voxposer} used pre-trained LLM and VLM to infer the affordances and constraints within the environment based on the language instructions provided. 
This approach enabled the execution of physical robot actions without relying on robotic data.
In our framework, we apply the pretrained VLM to align the visual information with semantic information in a similar way and use the LLM together with past experience as a reference to perform spatial reasoning in estimating the goal position in a zero-shot manner.


\section{Methodology}\label{methods}

The main objective of our framework (See Fig.~\ref{fig:framework}) is to leverage knowledge from previous arrangement experience and infer optimal positions for object placement conditioned on new instructions. First, the robot performs vision-language grounding by using semantic understanding provided by the Contrastive Language-Image Pre-Training Model (CLIP). This is supported by vision precision from a Segment Anything Model (SAM). It allows the robot to identify \textit{what} objects and \textit{where} they are in the environment. 
Then, we apply a Retrieval-Augmented Generation (RAG)~\cite{lewis2020retrieval} system to enable the robot to leverage past successful experiences for a rearrangement task.
Particularly, we apply the LLM to associate the most relevant experience with the given instruction and use the spatial and semantic information from this similar experience as a reference.
Next, we generate a prompt combining the spatial and semantic information of both the observed environment and the referenced experience. Based on this prompt, the LLM can accurately predict the position for object placement in rearrangement tasks, conditioned on the language instruction.

\subsection{Vision-language Grounding}
For language-conditioned rearrangement, vision-language grounding, which is the aligning of each object with the corresponding semantic information, plays a fundamental role in the task.
To predict the goal state of each object, we need to create an object-level representation, using the RGB observation of the environment from the camera.
Towards this, we apply a pre-trained SAM for segmenting objects in RGB images. The model uses images and point-based prompts as inputs and then produces the segmented object. 
By sampling single-point prompts across a grid on the image, SAM is able to generate a mask and minimum bounding box for each sampled location.
Concretely, given an input RGB image $\mathbf{I} \in \mathbb{R}^{H \times W \times 3}$, SAM segments object regions $\mathbf{M}_i$, each associated with a minimum bounding box $\mathbf{B}_i = \left(x_i, y_i, w_i, h_i, {\theta}_{i}\right)$, where $\left(x_i, y_i\right)$ denotes the centre point of the box, $\left(w_i, h_i\right)$ are the width and height of the box, and ${\theta}_{i}$ is the rotation of the rectangle. By using the SAM, we obtain the masked images and corresponding minimum bounding boxes: 
\begin{equation}
\mathbf{M}_i, \mathbf{B}_i=\operatorname{SAM}\left(\mathbf{I}, \mathbf{p}_i\right)
\end{equation}
where $p_i$ represents sampled points on the image in a grid.

After generating the masks across a grid, semantic filtering is performed on the image to obtain the desired objects in the language instruction.
In order to understand the objects to be manipulated, we first feed the LLM with language instruction and the prompt $\mathcal{Q}_{obj}$: \textit{``List the objects that are directly involved in the interaction described in the instruction"}. In response, the LLM identifies and outputs the objects related to the instruction. This is represented as $\mathcal{C}_{{obj}}=\operatorname{LLM}(\mathcal{L}, \mathcal{Q}_{obj})$ where $\mathcal{L}$ represents the given instruction.
For instance, if the given instruction is \textit{``put the apple next to the banana"}, the LLM will output the objects involved as [``apple", ``banana", ``others"] where the ``others" is given for the irrelevant objects.
To align segmented objects with semantic information, the open-vocabulary image classifier CLIP is used to match the masked images with corresponding names. 
Particularly, the visual features and semantic features are extracted by the visual encoder and the semantic encoders of CLIP respectively. We then compute the cosine similarity between the visual and text (i.e., object name) features:
\begin{equation}
s_{i, c}=\frac{\mathbf{v}_i \cdot \mathbf{t}_c}{\left\|\mathbf{v}_i\right\|\left\|\mathbf{t}_c\right\|}
\end{equation}
where $v_i$ and $t_c$ denote the visual features of segmented objects and semantic features respectively. Finally, we assign the object to the category with the highest similarity, which is denoted as $c_i=\arg \max _{c \in \mathcal{C}_{\text {obj }}} s_{i, c}$.
We can easily use the corresponding minimum bounding boxes and masked objects to obtain the centroid $m_i$ and rotation angle $r_i$ of each object. As a result, an object representation $
\mathcal{O}=\left\{\left(c_i, m_i, r_i, \mathbf{B}_i\right)\right\}_{i=1}^N
$ is generated, which includes spatial information such as centroid, bounding box, rotation, and the corresponding object names.

\subsection{Spatial reasoning with retrieval augmentation generation }
The key to object rearrangement is identifying the position where the target object should be placed.
Based on the spatial information from observed images, the LLM can be used to perform spatial reasoning to infer the objects' target position for the rearrangement task. 
\begin{figure}[t]
	\centering
	\includegraphics[width=0.95\columnwidth]{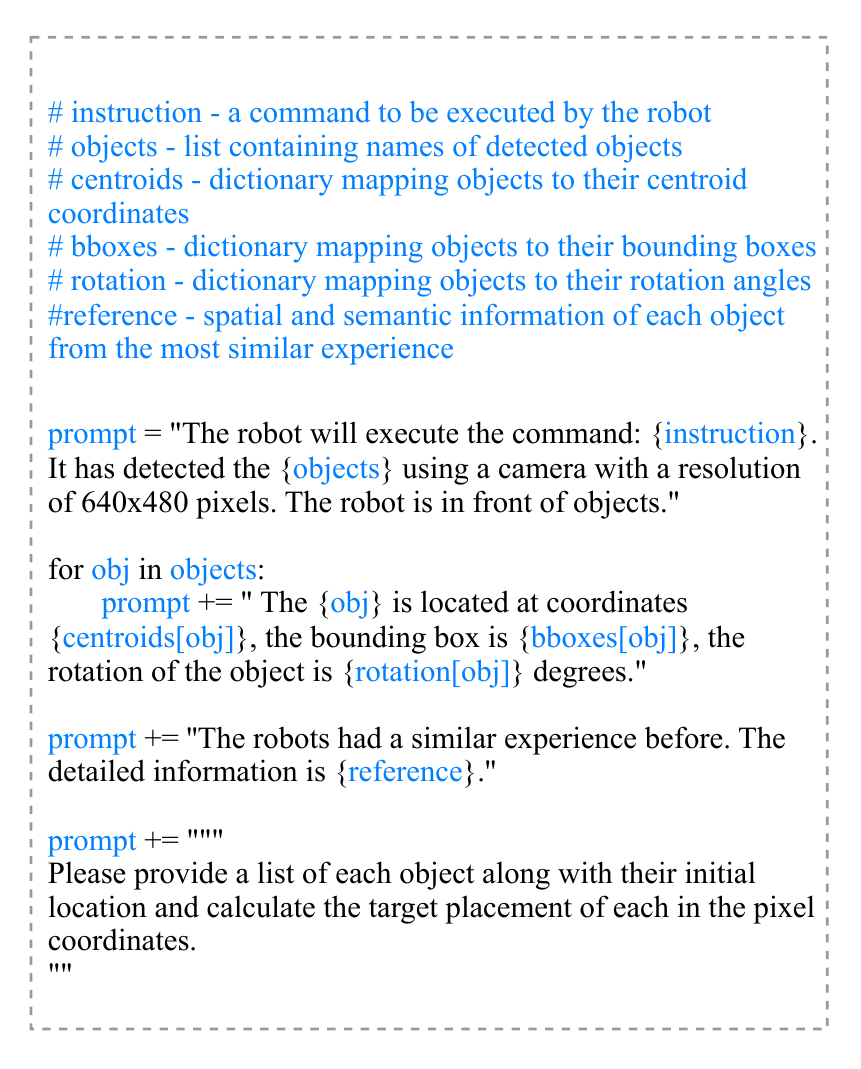}
	\caption{\textbf{\textit{Prompt engineering.}}  The simplified prompt for LLM to perform spatial reasoning. It includes the spatial and semantic information from both the observed RGB image and a similar successful experience.
 }
	\label{fig:prompt}
\vspace{-1.0em}
\end{figure}
However, the dataset used to train the LLM is limited, as many LLMs are trained on corpora from the internet, books, and videos, but lack spatial information. 
Consequently, LLMs often struggle with spatial reasoning abilities and can not understand the shape accurately~\cite{sharma2023exploring}.

Inspired by the Retrieval Augmentation Generation (RAG)~\cite{lewis2020retrieval}, we apply a RAG system that allows the LLM to retrieve ``outer" (that is not used in training) knowledge of past successful arrangement experiences and so enhance its generated results.
Given the outer knowledge of successful rearrangements $\mathcal{E}=\left\{\mathcal{E}_j\right\}_{j=1}^M$, where each experience $\mathcal{E}_j$ consists of a language instruction $\mathcal{L}_j$ and the object representation $\mathcal{O}_j$ which is arranged by humans.
The system identifies the most similar arrangement from the outer knowledge as a reference. 
As shown in Fig.~\ref{fig:framework}, the LLM scores the similarity between the new instruction and each instruction in past successful experiences.
Particularly, the similarity score can be denoted as ${s}_{\text {j }}=\operatorname{LLM}(\mathcal{L}, \mathcal{L}_{j}, \mathcal{Q}_{sim})$ and $\mathcal{Q}_{sim}$ represents the prompt: \textit{``Give a similarity score between two instructions on a scale from 0 to 100 "}.
The experience with the highest similarity score is selected as the reference: $\mathcal{E}^*=\arg \max _{\mathcal{E}_j \in \mathcal{E}} s_j$.
Detailed spatial and semantic information from the most closely matched experience is incorporated into the prompt to facilitate spatial reasoning.
An example of a simplified prompt template is shown in Fig.~\ref{fig:prompt}.
The LLM then predicts the target placement position $P_{t}$ as:
\begin{equation}
\mathbf{P}_t=\operatorname{LLM}\left(\mathcal{O}, \mathcal{L}, \mathcal{E}^*\right).
\end{equation}
Here, $\mathbf{P}_t=\left(x_t, y_t, r_t\right)$ represents the predicted coordinates and the rotation angle for object placement. 
Specifically, the LLM is designed to maintain the initial rotation angle to simplify the rearrangement, modifying it if a collision or overlap is detected. 

  
		

\subsection{Execute Pick and Place}

Once the LLM identifies the object to be moved and the desired position $\mathbf{P}_t=\left(x_t, y_t, r_t\right)$ for placement in the pixel coordinate, the robot can pick and place the object for the rearrangement. An RGB-D camera is mounted on the wrist of the robotic arm. Given the object’s initial and target locations in pixel coordinates, we can map these to their real-world coordinates using the depth information provided by the camera. The grasping pose is determined based on the object's minimum bounding box, where the gripper is aligned perpendicular to the longest edge of the object and centred to the object's centroid. We then employ inverse kinematics and a motion planner to find the path for the manipulation and move the object to complete the rearrangement task.

\begin{figure}[t]
	\centering
	\includegraphics[trim=0 0 0 -5, clip, width=0.9\columnwidth]{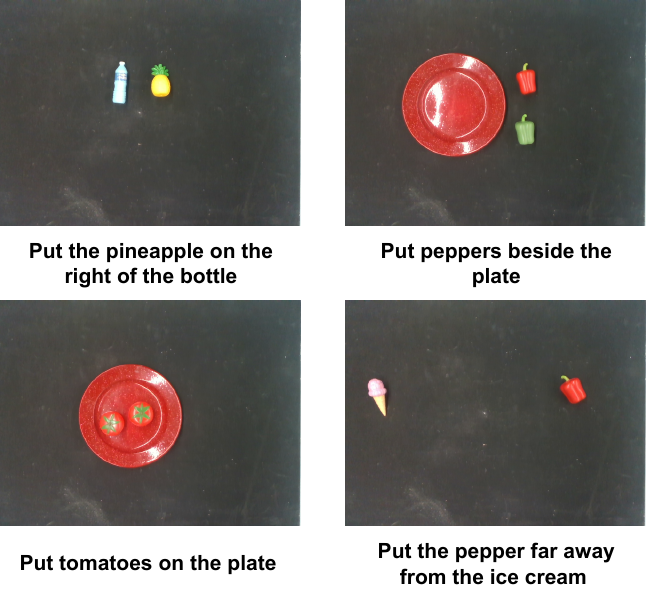}
	\caption{\textbf{\textit{Examples of successful rearrangements by humans.}} Four examples of successful rearrangements arranged by humans, with corresponding instructions.
 }
	\label{fig:example}
\vspace{-1.0em}
\end{figure}


\begin{figure*}
	\centering
	\includegraphics[trim=0 0 0 0, clip, width=1.75\columnwidth]{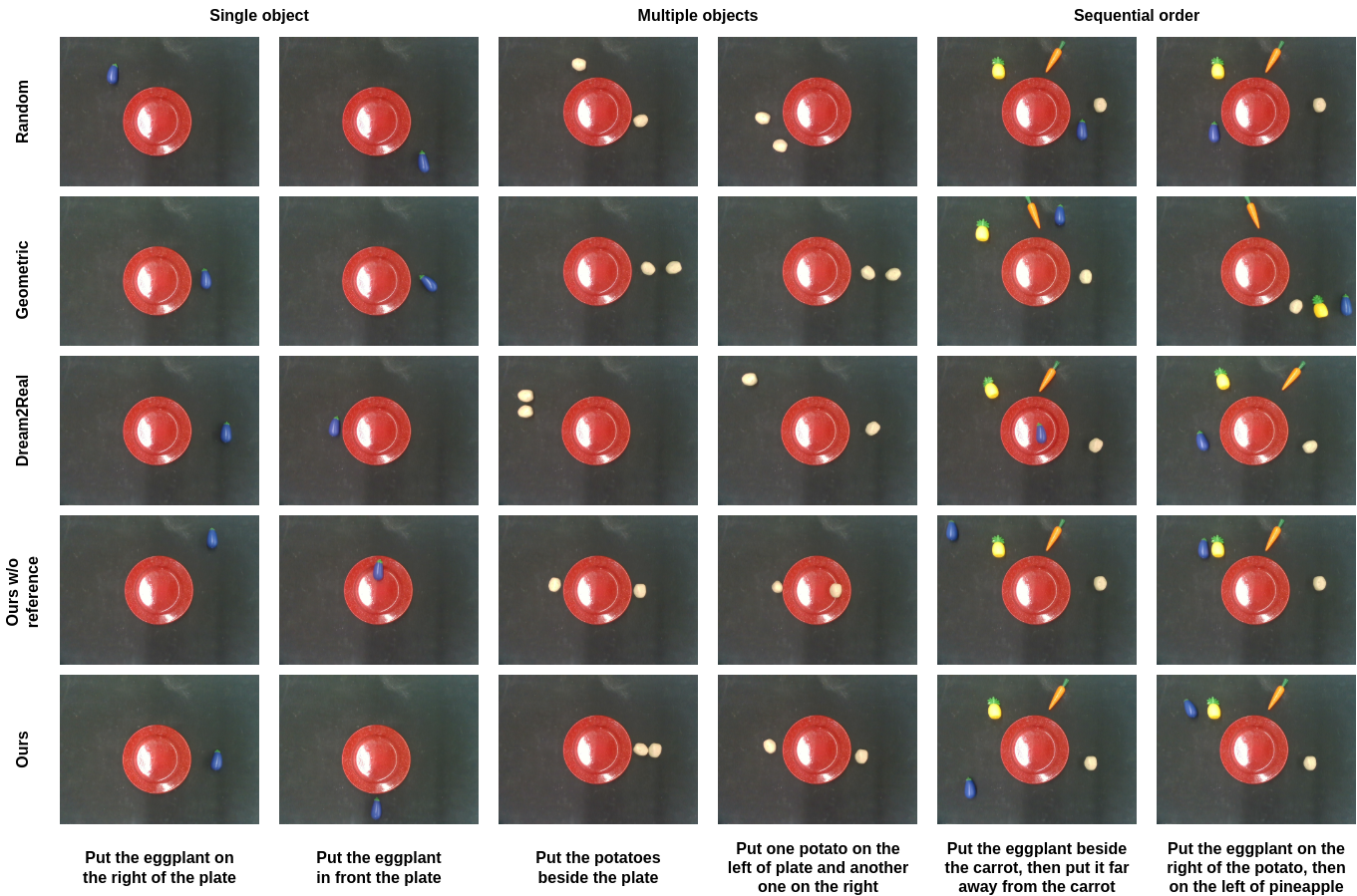}
	\caption{\textbf{Demonstration of results.}
The first row shows results from random object placement. The second displays objects arranged in a horizontal line with a certain gap. The third shows results from the application of Dream2Real~\cite{kapelyukh2024dream2real} while the fourth row is our proposed method without a reference or template (having a similar structure with TidyBot~\cite{wu2023tidybot}). The fifth row presents results using our complete proposed method with reference.
	} 
	\label{fig:visulisation}
 \vspace{-1 em}
\end{figure*}
\section{Experiments and analysis}\label{experiment}
\subsection{Robot Setup}
For our experiment, we apply a UR5e robotic arm, which is equipped with an OnRobot RG2 gripper. 
An Intel RealSense D435 RGB-D camera is attached to the wrist of the UR5e robotic arm to obtain the RGB image and depth information. 
The robot is controlled via the Robot Operating System (ROS) and we use the RRT-Connect algorithm for path planning.
\subsection{Outer Knowledge of Successful Rearrangement}
It is important to note that this outer knowledge of successful rearrangements is not used for training the model but rather serves as a reference to enhance the LLM's spatial reasoning capabilities.
The external knowledge from a successful experiment includes both spatial and semantic information, such as the name of each object, the object's centroid, its bounding box, rotation, and the corresponding instructions.
10 arrangements were created manually using representations of kitchen items.
Specifically, the image of the successful human-arranged object rearrangement is captured from a top-down view.
Based on the objects shown in the image, we use SAM to segment the corresponding masks using location prompts. From these masks, we derive the centroid, bounding box, and rotation of each object.
Eight of the scenarios contain two objects, while the remaining two contain multiple objects.
Some examples are shown in the Fig~\ref{fig:example}.
Our framework also has the potential to expand external knowledge by recording the spatial information of objects during successful rearrangements performed by robots. This can then be used as successful past experiences for future tasks similarly to Fig. \ref{Architecture}. 
However, in our experiments, we use only these 10 arrangements as outer knowledge to highlight how they improve the results.
More details are available on our webpage at: \url{https://sites.google.com/view/learnpast}

\subsection{User Study on Object Rearrangement}

In our experiment, 15 arrangements are performed on 3 real-world task scenarios.
The first scene consists of only an eggplant and a plate, with the robot manipulating a \textbf{single object}. The instructions for this scene are: (1) \textit{“put the eggplant on the right of the plate,”} (2) \textit{“put the eggplant on the left of the plate,”} (3) \textit{“put the eggplant in front of the plate,”} (4) \textit{“put the eggplant behind the plate,”} and (5) \textit{“put the eggplant far away from the plate.”} 
The second scenario include repetitive objects, i.e., two potatoes, to assess whether our method can effectively rearrange identical items and \textbf{multiple objects}. The instructions for the second scene are: (1) \textit{“put the potatoes on the plate,”} (2) \textit{“put the potatoes beside the plate,”} (3) \textit{“put one potato to the left of the plate and the other to the right,”} (4) \textit{“put the potatoes far away from the plate,”} (5) \textit{“put the potatoes together.”} 

In the third scene, our objective is to determine if our method can effectively rearrange objects in a long \textbf{sequential order}. This scene includes several objects, such as a carrot, a potato, a pineapple, an eggplant and a plate.
The instructions for the third scene are: (1) \textit{“put the eggplant on the plate, then beside the plate,”} (2) \textit{“put eggplant beside the plate, then beside the carrot,”} (3) \textit{“put the eggplant beside the potato, then put the eggplant on the plate,”} (4) \textit{“put the eggplant beside the carrot, then far away from the carrot,”} (5) \textit{“put the eggplant on the right of the potato, then on the left of the pineapple.”} 

For the evaluation of object rearrangement, our objective is to measure if the final arrangement of objects matches the given instructions. 
Since our framework aims to mimic human thinking and ensure that the rearrangement aligns with human preferences, a straightforward approach is to conduct a user study to gather human feedback, following the methodology in~\cite{sarch2022tidee} and~\cite{kapelyukh2023dall}.  
Specifically, 15 participants were invited to evaluate whether the final real-world scene created by the robot meets the given instructions. Four of the participants were female, while the others were male, with ages ranging from 25 to 33. Participants rated the results on a scale from 1 to 10, where 1 indicates \textit{``not acceptable"} and 10 indicates \textit{``highly acceptable"}. 
All users were presented with the same set of images and the set-up for every method was similar. The participants evaluated the results from 9 different methods (shown in Table~\ref{tab:compare} and Table~\ref{tab:backbone}), providing a total of 2025 ratings.

As shown in Table~\ref{tab:compare}, we present a comparison of our results with other baseline methods. Since most existing approaches rely on large training datasets, they are not directly comparable to our method in a zero-shot manner. To address this, two heuristic-based baselines are selected for comparison. In the \textit{Random} baseline, the robot picks objects and places them at random locations. 
In the \textit{Geometric} baseline, the robot arranges the relevant objects in a horizontal line with a certain gap, ensuring their centre points are aligned at the same level.
We also adapted Dream2Real~\cite{kapelyukh2024dream2real} to fit our tasks and use it as a baseline. Dream2Real uses sampling to place objects in various locations in order to create a scene aligned with the goal caption.
Additionally, we compare our results with an ablated structure that does not use referencing or templating for spatial reasoning. This follows a similar structure described in~\cite{wu2023tidybot}  
where the LLM is used for categorisation. However, we applied the LLM for spatial reasoning in this work.

As illustrated in Table~\ref{tab:compare}, our proposed method achieves the highest average evaluation score of 9.14, outperforming all other methods. 
The results also demonstrate superiority across all three evaluation scenarios: \textbf{single-object placement, multi-object arrangement, and sequential ordering,} indicating that the robot can effectively reason about spatial relationships and arrange objects in a reasonable place.
Notably, even when operating without reference guidance, our framework maintains competitive performance (7.93 mean score), outperforming the sampling-based approach~\cite{kapelyukh2024dream2real} and heuristic-based baselines, which demonstrates the inherent effectiveness of our LLM-driven architecture.
By incorporating references from successful experiences, our approach further enhances spatial accuracy, increasing the mean evaluation score by 1.21 compared to results without reference.
This suggests that our approach effectively leverages prior successful experiences to enhance knowledge transfer, facilitating more efficient and intelligent object rearrangement.
The results in Fig.~\ref{fig:visulisation} also illustrate that the use of reference enhances the accuracy of object placement and enables a more reasonable gap between objects. 
However, it is observed that the performance in single-object rearrangement is lower compared to multiple objects and sequential order tasks. This is likely because the instructions for single-object placement contain more precise spatial terms such as \textit{``in front of"}, \textit{``behind"}, \textit{``left of"}, and \textit{``right of"}, which pose a greater challenge for the LLM in accurately inferring object positions. In contrast, the other two tasks involve less precise spatial terms, such as \textit{``beside"} and \textit{``far away from"}, which are easier for the model to perform reasoning.

\begin{table}[t]
	\centering
	\caption{Experimental results by using different methods}
	\label{tab:compare}
    \scalebox{1}{
	\begin{tabular}{c|c|c|c|c}
		\hline
		\multirow{2}{*}{Methods} & \multirow{2}{*}{{\makecell[c]{Single\\object}}} & \multirow{2}{*}{{\makecell[c]{Multiple\\objects}}} & \multirow{2}{*}{{\makecell[c]{Sequential\\order}}} & \multirow{2}{*}{Mean}
		\\ & & & & \\
		\hhline{=|=|=|=|=}
		\multirow{2}{*}{Random} & \multirow{2}{*}{\centering 1.63$\pm{1.39}$} & \multirow{2}{*}{\centering 5.57$\pm{2.26}$} & \multirow{2}{*}{\centering 4.61$\pm{3.44}$} & \multirow{2}{*}{\centering 3.94} \\
		& & & & \\
        \hline
		\multirow{2}{*}{Geometric} & \multirow{2}{*}{\centering 2.92$\pm{3.48}$} & \multirow{2}{*}{\centering 4.65$\pm{3.19}$} & \multirow{2}{*}{\centering 4.80$\pm{4.08}$} & \multirow{2}{*}{\centering 4.12} \\
		& & & & \\
        \hline
		\multirow{2}{*}{{\makecell[c]{Dream2\\Real~\cite{kapelyukh2024dream2real}}}} & \multirow{2}{*}{\centering 6.16$\pm{3.24}$} & \multirow{2}{*}{\centering 6.95$\pm{2.27}$} & \multirow{2}{*}{\centering 7.33$\pm{2.77}$} & \multirow{2}{*}{\centering 6.81} \\
		& & & & \\
		\hline
		\multirow{2}{*}{{\makecell[c]{Ours w/o \\reference~\cite{wu2023tidybot}}}} & \multirow{2}{*}{\centering 5.51$\pm{3.71}$} & \multirow{2}{*}{\centering 8.92$\pm{1.85}$} & \multirow{2}{*}{\centering 9.37$\pm{1.17}$} & \multirow{2}{*}{\centering 7.93} \\
		& & & & \\
		\hline
		\multirow{2}{*}{Ours} & \multirow{2}{*}{\centering \textbf{8.67$\pm{\textbf{2.18}}$}} & \multirow{2}{*}{\centering \textbf{9.24$\pm{\textbf{1.44}}$}} & \multirow{2}{*}{\centering \textbf{9.51$\pm{\textbf{0.86}}$}} & \multirow{2}{*}{\centering \textbf{9.14}} 
		\\ & & & & \\
		\hline
	\end{tabular}}
	\vspace{-1.0em}
\end{table}

\subsection{Comparison between Different Language Models}
In this experiment, we apply various language models to evaluate how their capabilities affect the prediction of goal positions in object rearrangement. 
Instead of using an ``extra" Large Language Model like ChatGPT-4, we used smaller LLMs to assess whether our proposed framework can still achieve adequate performance with models that require lower computational resources and with results comparable to more resource-intensive models.

Specifically, we use different backbone models Mistral-7B~\cite{jiang2023mistral} and Llama3-8B~\cite{dubey2024llama} for spatial reasoning. These models are applied in two configurations: with and without reference.
From Table~\ref{tab:backbone}, we can observe that using a reference improves performance across all models in most tasks, with an increase of 1.21, 1.52 and 2.24 for ChatGPT4, Llamma3 and Mistral respectively.
In particular, the mistral model gains significant improvement from references in tasks involving multiple object rearrangement, with an increase of 3.78 in the score. 
It can be seen that ChatGPT-4 outperforms the smaller models across nearly all scenarios, regardless of whether a reference is used. 
However, the performance gap between ChatGPT-4 and the smaller models narrows when a reference is provided.
Given that ChatGPT4 is significantly larger than Mistral-7B and Llama3-8B, the result suggests that the use of references can mitigate the limitations caused by decreased computational capacity.

\begin{figure}[t]
	\centering
	\includegraphics[trim=60 260 80 260, clip, width=1\columnwidth]{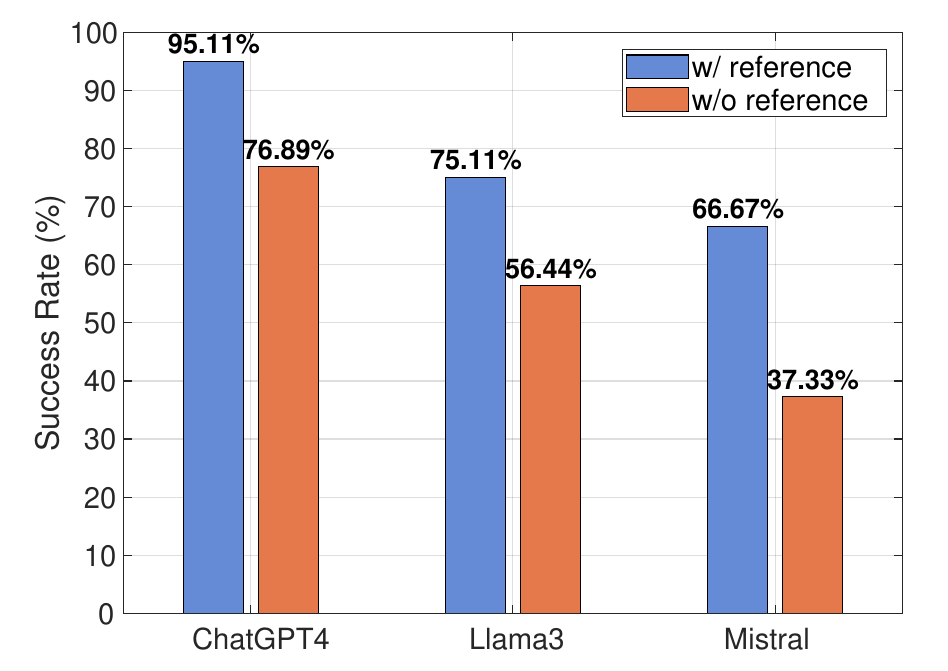}
	\caption{\textbf{\textit{Overall success rate.}} Success rates using different LLMs. The blue and orange bars represent the results with and without using references, respectively.
 }
	\label{fig:success}
\end{figure}

Additionally, to make the results more clearer, we present the success rate of rearrangement in Fig.\ref{fig:success}. 
We define a rearrangement as successful if its evaluation score is greater or equal to 7 and calculate the overall success rate as the number of ratings $\geq 7$ divided by the total number of ratings.
The results follow a similar trend where the ChatGPT using reference achieves the highest success rate of 95.11\%.
Additionally, it can be seen that incorporating reference guidance improves rearrangement performance by 18.22\%, 18.67\% and 29.34\% for ChatGPT-4, Llama3, and Mistral respectively. This demonstrates that reference is particularly beneficial for models with lower computational requirements.

\begin{table}[t]
	\centering
	\caption{Experimental results by using different backbones}
	\label{tab:backbone}
    \scalebox{1}{
	\begin{tabular}{c|c|c|c|c}
		\hline
		\multirow{2}{*}{Model} & \multirow{2}{*}{{\makecell[c]{Single\\objects}}} & \multirow{2}{*}{{\makecell[c]{Multiple\\objects}}} & \multirow{2}{*}{{\makecell[c]{Sequential\\order}}} & \multirow{2}{*}{Mean}
		\\ & & & & \\
		\hhline{=|=|=|=|=}
		\multirow{2}{*}{{\makecell[c]{Mistral\\w/o reference}}} & \multirow{2}{*}{\centering 3.69$\pm{3.42}$} & \multirow{2}{*}{\centering 3.21$\pm{2.75}$} & \multirow{2}{*}{\centering 7.69$\pm{2.94}$} & \multirow{2}{*}{\centering 4.87} \\
		& & & & \\
        \hline
		\multirow{2}{*}{{\makecell[c]{Llama3\\w/o reference}}} & \multirow{2}{*}{\centering 5.39$\pm{3.34}$} & \multirow{2}{*}{\centering 6.00$\pm{4.19}$} & \multirow{2}{*}{\centering 6.84$\pm{3.86}$} & \multirow{2}{*}{\centering 6.08} \\
		& & & & \\
		
            \hline
		\multirow{2}{*}{{\makecell[c]{ChatGPT4\\w/o reference}}} & \multirow{2}{*}{\centering 5.51$\pm{3.71}$} & \multirow{2}{*}{\centering 8.92$\pm{1.85}$} & \multirow{2}{*}{\centering 9.37$\pm{1.17}$} & \multirow{2}{*}{\centering 7.93} \\
		& & & & \\

  \hline
		\multirow{2}{*}{{\makecell[c]{Mistral\\w/ reference}}} & \multirow{2}{*}{\centering 5.24$\pm{3.70}$} & \multirow{2}{*}{\centering 6.99$\pm{3.37}$} & \multirow{2}{*}{\centering 9.09$\pm{1.32}$} & \multirow{2}{*}{\centering 7.11} \\
		& & & & \\
  \hline
		\multirow{2}{*}{{\makecell[c]{Llama3\\w/ reference}}} & \multirow{2}{*}{\centering 7.83$\pm{3.21}$} & \multirow{2}{*}{\centering 6.48$\pm{2.93}$} & \multirow{2}{*}{\centering 8.48$\pm{2.11}$} & \multirow{2}{*}{\centering 7.60} \\
		& & & & \\
	
        \hline
		\multirow{2}{*}{{\makecell[c]{ChatGPT4\\w/ reference}}} & \multirow{2}{*}{\centering \textbf{8.67$\pm{\textbf{2.18}}$}} & \multirow{2}{*}{\centering \textbf{9.24$\pm{\textbf{1.44}}$}} & \multirow{2}{*}{\centering \textbf{9.51$\pm{\textbf{0.86}}$}} & \multirow{2}{*}{\centering \textbf{9.14}} 
		\\ & & & & \\
		\hline
  
	\end{tabular}}
\end{table}

\section{Conclusion}\label{conclusion}
In order to achieve flexibility and efficiency, humans make use of past experiences for reasoning in novel task situations. On the other hand, most of the current methods in robotics make use of large datasets to train robots to achieve required goal states from a starting state, which reduces the flexibility and efficiency of robots in dealing with varying situations. 
To bridge this gap, this paper proposes a framework that leverages past successful rearrangements as references or templates to achieve efficient and flexible object manipulation, thereby mimicking human reasoning (see \cite{klein2022medial}\cite{fine2022whole}). This results in better prediction of target positions for accurate object placement.
Our results demonstrate that the use of references improve performance across multiple dimensions, including the rearrangement of a single object, multiple objects and tasks involving instructions with a sequential order. Additionally, the application of outer knowledge can help overcome the limitations caused by computational constraints, making it attractive for local deployment in practical applications. 
Nevertheless, the proposed method has some limitations that may affect its performance in real-world applications. Firstly, the scenario we designed operates primarily on a 2D surface, rather than in a 3D space, which limits its applicability in the physical world. Additionally, the objects used in the space are relatively sparse and this could pose a challenge in more clustered environments. Future improvements would involve extending the method to 3D rearrangement and addressing object rearrangement in clustered settings.


{\small
	\bibliographystyle{ieeetr}
	\bibliography{reference.bib}
}

\end{document}